%% file: main.tex
\documentclass[letterpaper, 10 pt, conference]{IEEEtran}
\IEEEoverridecommandlockouts

\input{header}

\begin{document}

    \title{
        Mission Design for Unmanned Aerial Vehicles  \\
        using Hybrid Probabilistic Logic Programs
    }
    
    \author{
        Simon Kohaut$^{1}$, Benedict Flade$^{2}$, Devendra Singh Dhami$^{1, 3}$, Julian Eggert$^{2}$,  Kristian Kersting$^{1, 3, 4, 5}$
        \thanks{
            $^{1}$ Artificial Intelligence and Machine Learning Lab, \newline\hspace*{1.6em} 
            Department of Computer Science, \newline\hspace*{1.6em}
            TU Darmstadt, 64283 Darmstadt, Germany \newline\hspace*{1.6em}
            {\tt\small firstname.surname@cs.tu-darmstadt.de}%
        }%
        \thanks{
            $^{2}$ Honda Research Institute Europe GmbH, \newline\hspace*{1.6em} 
            Carl-Legien-Str. 30, 63073 Offenbach, Germany \newline\hspace*{1.6em}
            {\tt\small firstname.surname@honda-ri.de}
        }%
        \thanks{
            $^{3}$ Hessian AI
        }%
        \thanks{
            $^{4}$ Centre for Cognitive Science
        }%
        \thanks{
            $^{5}$ German Center for Artificial Intelligence (DFKI)
        }%
    }
    
    \maketitle

    \begin{abstract}
        \input{content/0_abstract}
    \end{abstract}

    \input{content/1_introduction}
    \input{content/2_related_work}
    \input{content/3_methods}
    \input{content/4_experiments}

    \input{content/5_conclusion}
    \input{content/6_acknowledgment}

    \bibliographystyle{IEEEtran}
    \bibliography{main.bib}
\end{document}

%% file: header.tex
\usepackage{cite}
\usepackage{amsmath,amssymb,amsfonts}
\usepackage{algorithmic}
\usepackage{graphicx}
\usepackage{textcomp}
\usepackage[dvipsnames]{xcolor}
\usepackage{balance}
\usepackage{epstopdf}
\usepackage[normalem]{ulem}
\usepackage{upgreek}
\usepackage{comment}
\usepackage{url}
\usepackage{tikz}
\usepackage{lipsum}
\usepackage[formats]{listings}
\usepackage{subcaption}
\usepackage{multirow}
\usepackage[frozencache,cachedir=.]{minted}
\usepackage{tikz}
\usepackage[color=black,opacity=1,angle=0,scale=1]{background}
\backgroundsetup{
  contents={
    \begin{tikzpicture}
        \node at (current page.center) [align=center] {\textcopyright 2023 IEEE. Personal use of this material is permitted. \\ Permission from IEEE must be obtained for all other uses, in any current or future media, including reprinting/republishing this material for advertising or promotional purposes, \\ creating new collective works, for resale or redistribution to servers or lists, or reuse of any copyrighted component of this work in other works.
         \\ DOI: 10.1109/ITSC57777.2023.10422083};
    \end{tikzpicture}},
  placement=bottom,
  scale=0.6,
  vshift=20
}

%% file: content/0_abstract.tex
Advanced Air Mobility (AAM) is a growing field that demands a deep understanding of legal, spatial and temporal concepts in navigation.
Hence, any implementation of AAM is forced to deal with the inherent uncertainties of human-inhabited spaces. 
Enabling growth and innovation requires the creation of a system for safe and robust mission design, i.e., the way we formalize intentions and decide their execution as trajectories for the Unmanned Aerial Vehicle (UAV).
Although legal frameworks have emerged to govern urban air spaces, their full integration into the decision process of autonomous agents and operators remains an open task.
In this work we present ProMis, a system architecture for probabilistic mission design.
It links the data available from various static and dynamic data sources with legal text and operator requirements by following principles of formal verification and probabilistic modeling.
Hereby, ProMis enables the combination of low-level perception and high-level rules in AAM to infer validity over the UAV's state-space.
To this end, we employ Hybrid Probabilistic Logic Programs (HPLP) as a unifying, intermediate representation between perception and action-taking.
Furthermore, we present methods to connect ProMis with crowd-sourced map data by generating HPLP atoms that represent spatial relations in a probabilistic fashion.
Our claims of the utility and generality of ProMis are supported by experiments on a diverse set of scenarios and a discussion of the computational demands associated with probabilistic missions.


%% file: content/1_introduction.tex
\section{Introduction}
\label{sec:introduction}

\PARstart{M}{ission} design for intelligent, mobile agents presents several challenges and demands strict adherence to safety and security requirements. 
It involves defining the mission objectives, constraints, and execution strategies for navigating an Unmanned Aerial Vehicle (UAV). 
A valid mission consists of a sequence of trajectories leading to the successful fulfillment of the objectives without violating any operational, manufacturing, or public regulations.
To design a mission, tasks such as mission clearance, analysis and optimization need to be performed. 
While most research in mobility-related mission design focuses on optimizing the agent's trajectory \cite{chenMultiDrone, guanEfficientUAV, hohmann2022multi}, we believe that using declarative, relational models can provide a more holistic and symbolic view of the navigation problem. 
Our approach aims to improve UAV navigation by providing a system that leverages, among other data sources, crowd-sourced maps to provide insights into the agent's freedom of movement within a legal framework modelled by probabilistic logic.
For example, Figure~\ref{fig:pml_matrix} shows how our system enables highly adaptable control over the UAV's navigation space. 
In summary, we seek to improve the mission design for UAVs at a fundamental level, by leveraging probabilistic logic to increase the robustness, efficiency and safety of intelligent vehicles navigating complex spaces.

To make way for future assistance in service applications and innovative logistics networks, it is critical to expand the areas in which we can safely deploy autonomous agents.
Since these applications often operate in public spaces or directly interact with humans, it is crucial to carefully define the limits of the agent's action and task spaces.
In the context of Advanced Air Mobility (AAM), we face numerous challenges related to the use of UAVs:

\input{figures/pml_matrix}

\begin{itemize}
    \item \textbf{Safety}:
    Ensuring reliable operations of UAVs is a multi-faceted task that involves the safety of both ground and airborne entities.
    It requires a consideration of various issues, including collision avoidance, weather conditions, or hardware malfunctions.
    Especially keeping safe \textit{distances} to buildings and infrastructure or not flying \textit{over} the heads of bystanders are common requirements.
    \item \textbf{Regulations}:
    The use of drones for private and commercial purposes is subject to numerous regulatory barriers. 
    Some of these regulations are contingent on the environment of the UAV during the mission, such as the movement of bystanders and other traffic participants. 
    Utilizing the UAV's sensory equipment helps in this context to establish compliance with regulations and to allow for safe and effective operations.
    \item \textbf{Public perception}:
    Achieving widespread acceptance of drone services requires addressing public concerns regarding the use of UAVs. 
    This includes privacy infringement, noise pollution, and the potential for accidents. 
    The public is hesitant with accepting the use of uninterpretable black-box models to make critical decisions, since they pose a challenge to understand and evaluate the decision-making process.
\end{itemize}

\input{listings/problog_example}

These are reasons motivating the use of symbolic, white-box models for mission design.
More specifically, to represent the laws, relationships and uncertainties of the agent's navigation, probabilistic logic programs can be a suitable fit.
Logic Programming allows for programmatic reasoning within First-Order Logic (FOL).
On top of this, probabilistic extensions of Logic Programming exist to assign categorical distributions to the FOL model, i.e., its relations over the contained objects.
Listing~\ref{listing:problog_example} shows an exemplary model written in ProbLog \cite{problog}, one of the most commonly used variants of probabilistic logic programming today.
Hybrid Probabilistic Logic Programs (HPLP) make simultaneous use of discrete and continuous distributions, thereby representing a so-called hybrid relational domain where both probabilities and densities are applicable to the FOL \cite{nitti, kumar2022first}.
Especially in probabilistic robotics \cite{thrun2002probabilistic} where it is not enough to deal with categorical data, HPLPs are a promising research direction.

In this work, we combine Geographic Information Systems (GIS) with HPLPs by aligning state-of-the-art GIS typing systems, e.g., as in Listing~\ref{listing:overpass_ql}, with the probabilistic logic programs we generate within ProMis (see Section~\ref{sec:methods}).
Hence, with ProMis we present a hybrid probabilistic mission design architecture for AAM that uses HPLPs to model legal and safety constraints.
ProMis further employs probabilistic inference to facilitate UAV operations and eases the determination of safe and legal flight trajectories.
An open-source software implementation, enabling others to reproduce our results and further advance research in this area, is under development.

%% file: figures/pml_matrix.tex
\begin{figure}
    \centering
    \begin{subfigure}{0.39\linewidth}
        \includegraphics[width=\textwidth]{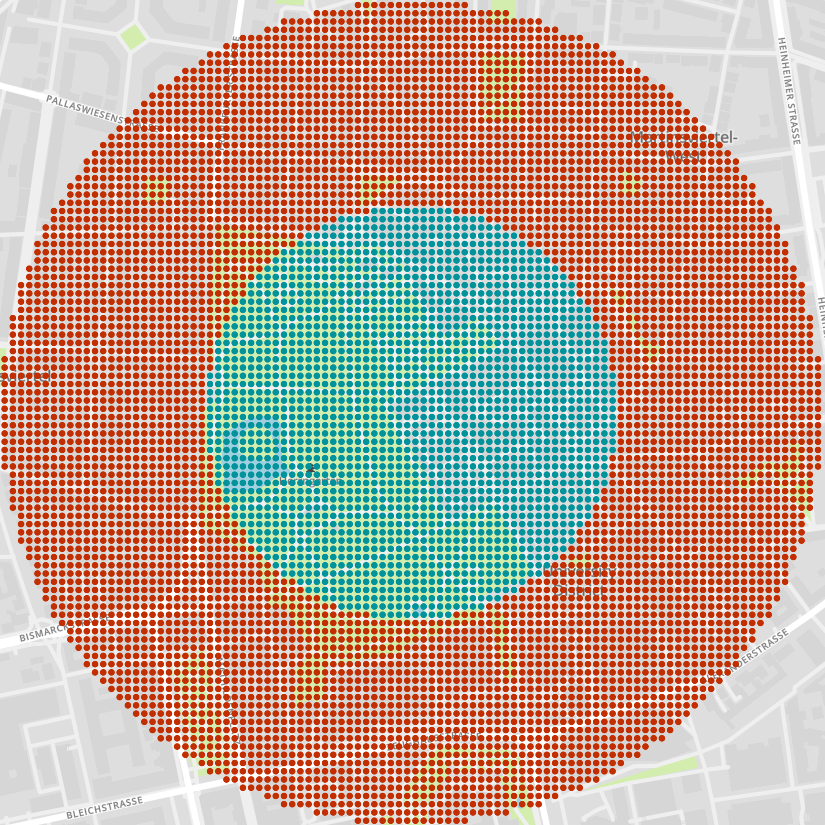}
    \end{subfigure} 
    \begin{subfigure}{0.39\linewidth}
        \includegraphics[width=\textwidth]{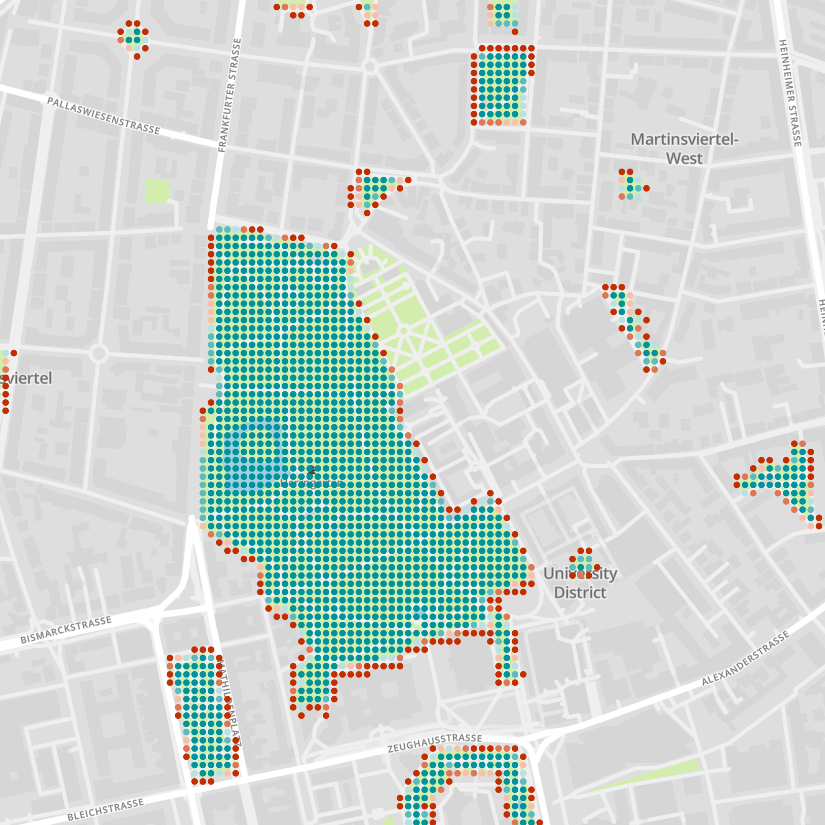}
    \end{subfigure} \\
    \vspace{0.35em}
    \begin{subfigure}{0.39\linewidth}
        \includegraphics[width=\textwidth]{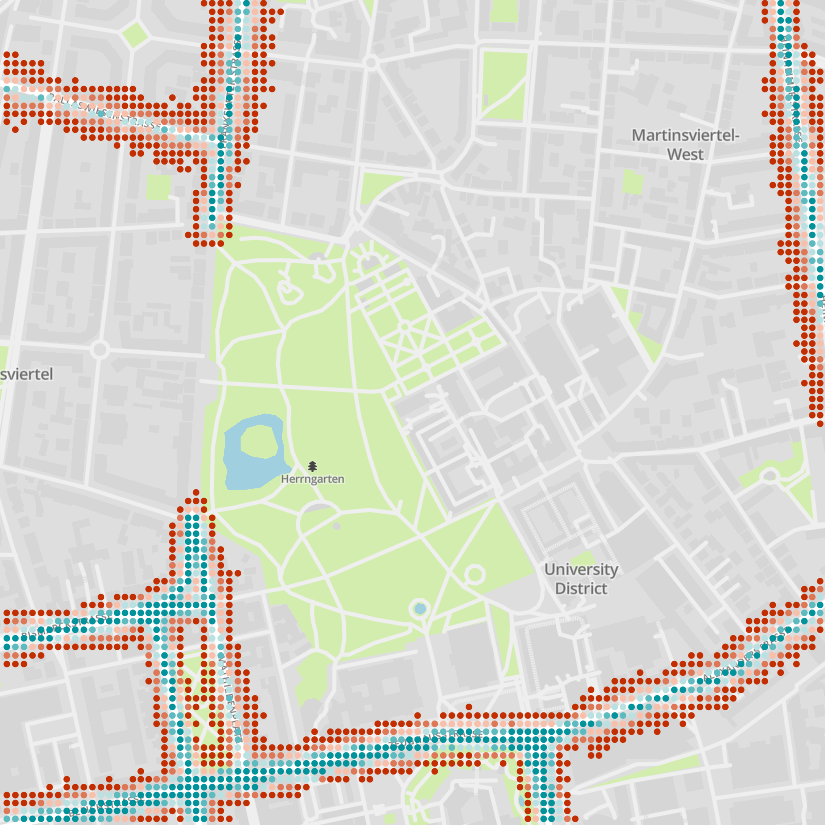}
    \end{subfigure} 
    \begin{subfigure}{0.39\linewidth}
        \includegraphics[width=\textwidth]{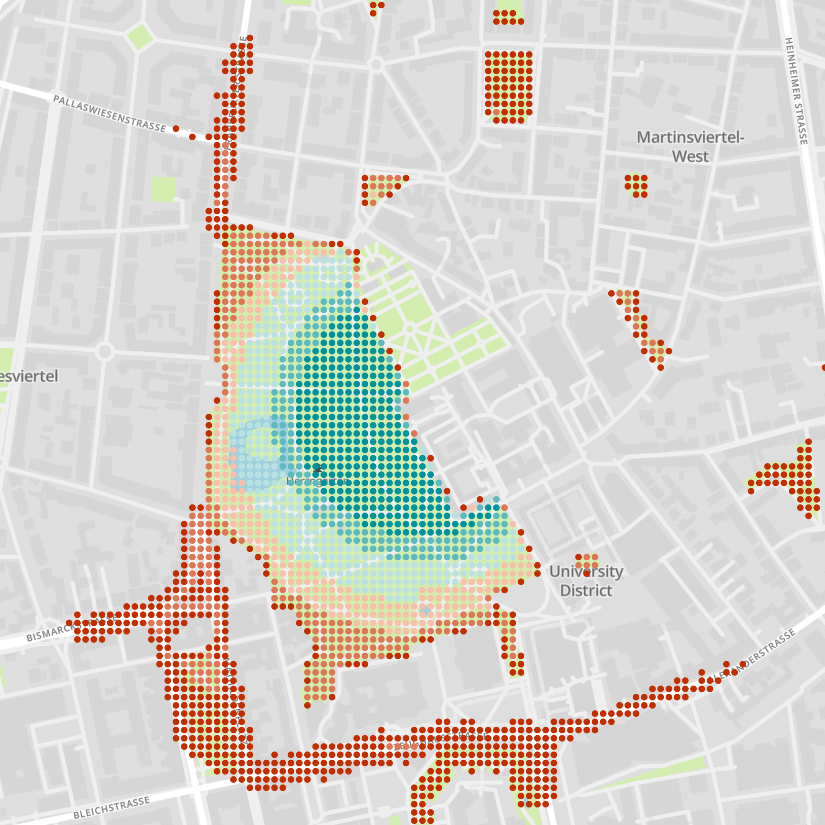}
    \end{subfigure} 
    \caption{
        \textbf{Adaptable mission design with ProMis:} 
        Probabilistic Mission Landscapes (PML) express the probability of a First-Order Logic model being satisfied over the navigation area.
        Each picture shows a different constraint and, on the bottom-right, a conjunction of all others.
        Cyan locations indicate a high probability of satisfying all constraints.
    }
    \label{fig:pml_matrix}
\end{figure}

%% file: listings/problog_example.tex
\begin{listing}[t]
    \centering
    \begin{minted}
    [
        frame=lines,
        autogobble,
        fontsize=\footnotesize
    ]{prolog}
        % A person that owns a drone is likely to operate it
        0.9::operates_drone(X) :- person(X), owns_drone(X). 
        
        % Friends might influence to also buy a drone 
        0.2::owns_drone(X) :- friend(X, Y), owns_drone(Y).
        
        % Background knowledge
        person(justus). 
        person(jonas).
        owns_drone(justus).
        friend(jonas, justus).
        
        % Chance of Jonas operating a drone
        query(operates_drone(jonas)).  % 0.18
    \end{minted}
    \caption{
        An example of probabilistic modelling in ProbLog \cite{problog}, a probabilistic programming language for discrete domains.
        This model describes how friends will influence each other in the decision of owning and operating a drone.
        Rules are annotated with probabilities, i.e., the uncertainty associated with the logical relations over the data.
    }
    \label{listing:problog_example}
\end{listing}

%% file: content/2_related_work.tex
\input{listings/overpass_ql}
\input{figures/architecture}

\section{Related Work}
\label{sec:related_work}

In Europe, the issue of unmanned aircraft systems and urban air mobility is driven by the Single European Sky Air Traffic Management Research (SESAR) initiative. 
When SESAR presented its first Master Plan in 2009~\cite{sesar_2009}, AAM was not considered and drones were only mentioned in passing.
With the publication of the 3rd edition of the Master Plan in 2015~\cite{undertaking2016european} and the "Drone Outlook Study" in 2017~\cite{sesar_2017}, the push for urban air mobility in Europe has been given green light. 
Compared to the established ecosystems for general aviation, the development of novel unmanned traffic management systems needs to address issues such as significantly higher traffic densities, gaining additional societal acceptance and clearing up uncertainties in regulations.
Therefore, a focus of this paper is put on the application of declarative probabilistic models for mission design and their feasibility in aviation.

To act in an environment full of static, semi-static and dynamic entities, a representation of the environment is required.
Particularly where a world view can be augmented with information about signals and dynamics, e.g., traffic lights and signs in autonomous driving, hierarchical approaches to mapping have emerged.
In particular, so-called local dynamic maps have become popular among autonomous driving researchers to describe lanes, traffic rules, congestion, and even individual road users~\cite{rldm, enriched_ldm,ldm_safety}.
As maps become more feature rich, the development of more advanced techniques for navigating public spaces becomes possible.
For example, matching local views of the environment can help an agent to locate itself on the road network~\cite{lane_detection, visual_map, urban_camera}.
Of course, using statistical methods for matching and decision-making can be a dangerous game, as mentioned above.
Even if the human is in full control, low situation awareness may quickly lead to a life-threatening scenario.
Therefore, keeping the operator or pilot informed of the ongoing mission is critical. 
For example, controlled flight into terrain accounts for a significant proportion of aviation fatalities~\cite{cfit_aviation}.
For this reason, rich visualizations of the mission in progress have been developed to avoid blind spots in decision-making~\cite{visualization_aviation}.
Inspired by air traffic management, it aims to improve flight safety by enabling operators to gain a better situational understanding and awareness of obstacles by displaying the intended trajectory and environment.
By providing insight into the mission and its environment from information sources, operators can react to threats and optimize their behavior~\cite{gis_aviation, weather_information_impact}.

To employ autonomous agents, beyond the necessity of a fine-tuned control system, a good understanding of the individual agent's design goals and application domain is crucial.
For instance, in agricultural applications, robots can leverage weather forecasts to plan their missions, enabling them to determine when and how to perform tasks in order to, e.g., avoid wet or muddy terrain~\cite{agriculture}. 
For an adaptable and interpretable system, symbolic reasoning systems are a suitable basis to formalize and verify behavior and its constraints.
One of the earliest programmatic reasoning systems in First-Order Logic (FOL) is Prolog~\cite{colmerauer1990introduction}, developed in 1972 by Alain Colmerauer.
Prolog has inspired numerous applications, from natural language processing~\cite{nitti} to robotics~\cite{robot_prolog}.
Extensions that embrace uncertainties in formal logic like Bayesian Logic Programs~\cite{bayesian_logic} and Probabilistic Logic Programs~\cite{problog,inference_in_plp} have been introduced to allow for probabilistic inference in FOL models.
While they are not formulated for end-to-end learning in tandem with artificial neural networks, newer models like DeepProblog~\cite{deepproblog} and SLASH~\cite{slash} have been introduced to close this gap.
The objective of such Neuro-Symbolic AI methods is to combine the strengths of deep learning architectures with symbolic, formally well justified and inherently interpretable systems.
Naturally, applying these methods in the context of building autonomous agents, e.g., robots acting in logically constrained environments, is interesting to ensure safe and reliable decision-making.
Our work in probabilistic mission design for AAM is motivated by hybrid formulations of probabilistic logic programming on which noteworthy progress has been made in recent years~\cite{nitti, kumar2022first}.
They extended the syntax and semantics of probabilistic logic programs to efficiently account for hybrid relational domains, i.e., domains in which both discrete and continuous distributions are required to model the agent's domain. 

%% file: listings/overpass_ql.tex
\begin{listing}[t]
    \centering
    \begin{minted}
    [
        frame=lines,
        autogobble,
        fontsize=\footnotesize
    ]{c}
    // Output format and timeout for request
    [out:json][timeout:25];

    // Requested ways within a bounding box
    // Here, we retrieve typed road-segments
    (
        way["highway"="primary"]({{bbox}});
        way["highway"="secondary"]({{bbox}});
        way["highway"="tertiary"]({{bbox}});
        way["highway"="footway"]({{bbox}});
    );

    // Return retrieved data
    out body; >; out skel qt;
    \end{minted}
    \caption{
        A common example for querying data from OpenStreetMap using the Overpass Query Language.
        Similarly to ProbLog, Overpass uses relational information that describes in its case the types of nodes, ways and areas.
        This script gives accesses to different types of streets that are related by the "highway" predicate.
    }
    \label{listing:overpass_ql}
\end{listing}

%% file: figures/architecture.tex
\begin{figure*}
    \centering
    \includegraphics[width=\textwidth]{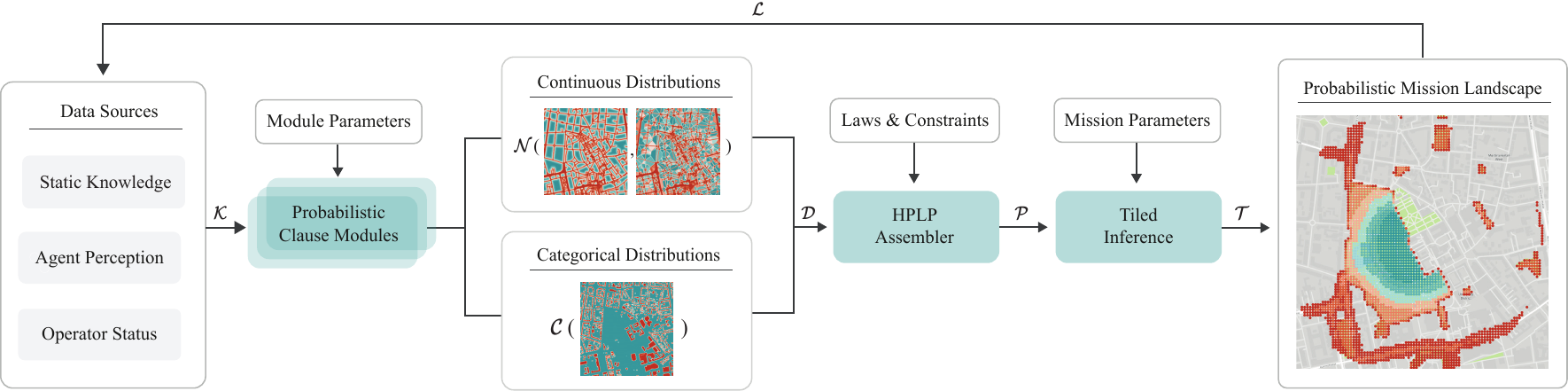}
    \caption{
        \textbf{The Probabilistic Mission (ProMis) system architecture:}
        Static and dynamic data sources are transformed by the Probabilistic Clause Modules (PCM) into continuous and categorical distributions.
        Then, the Hybrid Probabilistic Logic Program (HPLP) Assembler writes the retrieved location dependent parameters as distributional clauses and combines the hereby formed HPLP with predefined laws and constraints.
        Given the parameters of the mission, tiled inference over the agent's navigation space results in a Probabilistic Mission Landscape (PML) that can then be looped back to be stored, used for action planning or as advanced mission analysis tool for the operator. 
    }
    \label{fig:architecture}
\end{figure*}

%% file: content/3_methods.tex
\input{figures/distributional_atoms}

\section{Methods}
\label{sec:methods}

\subsection{An Architecture for Probabilistic Mission Design}

In this section we introduce an architecture for logically constrained, probabilistic mission design.
With ProMis, we aim at a \underline{Pro}babilistic \underline{Mis}sion design flow that addresses the needs of all involved stakeholders (see Figure~\ref{fig:architecture}).

In AAM, the following three parties represent the main stakeholders:
\begin{itemize}
    \item \textbf{The public body} defines and regulates public traffic areas.
    This includes the classifications and categorizations of autonomous agents and their operations, which gives rise to a need for developing safe and legally compliant mission models. 
    In addition to creating rules for public spaces, they provide data on critical infrastructure or areas that privately operated agents should not enter, e.g., geofences protecting the space around airports.
    For the most part, the public body thereby provides \textit{static knowledge} and the aforementioned \textit{laws}.
    \item \textbf{The operator} of the autonomous agent determines the mission's target and additional \textit{constraints} on top of legal requirements.
    While the operator's preferences cannot supersede those of the public body with regard to public grounds, they can impose additional restrictions on valid missions by specifying risk-averse rules or preferences.
    As operators may not possess the necessary engineering expertise to manually program their requirements, intuitive interfaces are necessary. 
    Additionally, operators with AAM specific qualifications have direct influence on what the autonomous agent is authorized to do under their supervision.
    \item \textbf{The manufacturer} has the most extensive knowledge of the autonomous agent's sensory equipment and algorithms for detection, localization, and tracking. 
    Their expertise should be integrated into the mission design system in the form of the \textit{agent's perception}.
    For example, a heavy drone may provide estimates of its weight and the movement of people below to avoid flying over a crowd of bystanders. 
    Manufacturers may also impose limitations on a vehicles use for licensing, safety, or hardware protection reasons.
    Therefore, they can further add to the \textit{constraints} of the mission.
\end{itemize}
ProMis incorporates the contributions of these three in the form of \textit{static knowledge}, \textit{laws and constraints} and \textit{agent perception}.
By extracting the parameters of distributions modelling the mission environment and combining with public and operator \textit{laws and constraints}, a HPLP for probabilistic inference in the mission design domain is constructed.
Our main contributions are (i) the architecture of ProMis, integrating uncertain data of AAM settings with probabilistic logic, (ii) the formalization of distributional clauses from typed map data and (iii) the creation of Probabilistic Mission Landscapes (PML).
PMLs express the validity of points in the UAV's state-space in the form of a scalar-field of probabilities.

\subsection{Hybrid Relational Navigation}

While HPLPs are domain agnostic models in which probabilistic logic can be utilized to query for arbitrary information, in this section we will focus on the treatment of spatial data for UAV navigation.
From the GIS, but also potentially other static and dynamic data sources like the agent or operator, we obtain background knowledge $\mathcal{K}$ of three types:
facts, estimates and geographic features.
While facts and estimates represent categorical and continuous data respectively, geographic features are for example line segments and areas in a polar, earth centered coordinate system.

We propose Probabilistic Clause Modules (PCM) as a collection of mappings from application-specific data models in $\mathcal{K}$ into a set of continuous and categorical distributions $\mathcal{D}$.
Hence, each entry in this distributional knowledge base $\mathcal{D}$ is a pair of an atom's name $\mathcal{A}$ and either its probability $p$ or its parametric distribution $P(\vec{\theta})$.

In other words, one of the following can be used to express the fact that $\mathcal{A}$ is true with probability $p$ or distributed according to $P(\vec{\theta})$.
So, we either obtain categorically or continuously distributed clauses:
\begin{align*}
    p &:: \mathcal{A} \leftarrow l_1, \ldots, l_n. \tag*{(Discrete)} \\
    \mathcal{A} &\sim P(\vec{\theta}) \leftarrow l_1, \ldots, l_n. \tag*{(Continuous)}
    \label{eq:distributional_clauses}
\end{align*}%
Such distributional clauses express that in all worlds where the conjunction of the \textit{literals} $l_1, \ldots, l_n$ is true, \textit{atom} $\mathcal{A}$ being true has probability $p$ or is distributed according to $P(\vec{\theta})$.
Note that $P(\vec{\theta})$ is not necessarily a distribution over probabilities, therefore $\mathcal{A}$ can take values from arbitrary domains.

\input{listings/spatial_relations}

While facts and estimates can be directly written in HPLP form, GIS data is an example where special treatment is necessary. 
To work within the spatial context of GIS-provided data and represent a basis for common laws and constraints in navigation, we introduce two PCMs for the binary relations $\textit{distance}(\textit{lat}, \textit{lon}, F)$ and $\textit{over}(\textit{lat}, \textit{lon}, F)$ respectively.
Here, $\textit{lat}$ and $\textit{lon}$ reference the latitude and longitude in an earth-centered navigation space and $F$ a set of environment features to be related to.

Since the creation and maintenance of high-definition maps is an intractable and expensive task, crowd-sourced geographic data is a promising substitution.
Unfortunately, such data is rarely accurate enough to allow for navigation without further safety considerations.
To express this spatial uncertainty, we apply a stochastic model to randomly sample each feature.
Here, we will consider two types of errors.
On one hand, map features can be simply translated, i.e., each vertex is offset by the same amount.
On the other hand, a feature can be rotated, sheared or scaled.
Hence, similar to prior work \cite{flade2021error}, we consider for each $f_{i, j} \in \mathbb{R}^2$, being the $j$-th vertex of the $i$-th feature, the following affine map: 
\begin{align*}
    \vec{\Phi}^n &\leftarrow \phi(\vec{\alpha}_i) \tag*{(Transformation)} \\
    \vec{t}^n &\leftarrow \kappa(\vec{\beta}_i) \tag*{(Translation)} \\
    \vec{f}^n_{i,j} &= \Phi^n \cdot \vec{f}_{i, j} + \vec{t}^n \tag*{(Generation)}
\end{align*}
This is done for each $n \in \{1, \ldots, N\}$ to obtain a collection of $N$ samples.
Here, the $\phi(\vec{\alpha}_i)$ generates matrices $\vec{\Phi}^n \in \mathbb{R}^{2 \times 2}$ to apply geometric transformations, e.g., rotations, that keep the center point fixed while $\kappa_{i, j}$ generates offset vectors $\vec{t}^n \in \mathbb{R}^2$ to apply a translation.
In this context $\vec{\alpha}_i$ and $\vec{\beta}_i$ are the parameters of the error per feature and define the expected offsets or geometric transformations.
Having defined this generator scheme, we can move on to the next step of computing the location dependent parameters of \textit{distance} and \textit{over}.

\input{listings/uam_model}

For the first PCM, \textit{distance}, we compute the distance of each location in navigation space to the nearest geometry of the relevant type of environmental feature.
The second PCM, \textit{over}, represents the occupancy of a feature set at a location in navigation space.
Therefore, this relation can be modeled as a scalar field of probabilities, indicating the chance of the feature being at that point.
In other words, we analyze how often each location is occupied in the typed sets of feature samples.
The parameters of the respective distribution are in both cases obtained via moment matching to the collection of $N$ \textit{distance} and \textit{over} samples.
Parameters of both distributional clauses can be seen in Figure~\ref{fig:distributional_atoms} when choosing to represent the \textit{distance} to roads using normal distributions and the \textit{over} relation of buildings in the environment.

\input{figures/scenarios}

\subsection{Probabilistic Mission Landscapes}

We now introduce the concept of Probabilistic Mission Landscapes (PML).
Given the mission parameters, constraints and inference parameters as described before, ProMis produces a grid of probabilities for a legal and safe arrival at each location.
For this, the earth-referenced (\textit{lat}, \textit{lon}) data is projected into a local, Cartesian space and discretized into a finite grid of given east- and north-resolution.
Instead of representing the previously introduced spatial relations in dependence of continuous latitude and longitude, we will continue with $\textit{distance}(r, c, F)$ and $\textit{over}(r, c, F)$.
All data assigned to the same row $r$ and column $c$ is aggregated and referenced in this grid.
Listing~\ref{listing:spatial_relations} demonstrates an excerpt of the output of these modules.

By defining a threshold for the probability to distinguish between valid and invalid missions across the landscape, operators and agents can determine where the next part of the operation can potentially lead. 
As a result, a single-trajectory validation is lifted from the mission clearance field to the scope of mission analysis and optimization. 
Operators can adjust their preferences to reflect a growing or shrinking landscape, while agents can plan a path within the landscape to stay within legal and operator boundaries.

Since the PML is based on geographic propositions such as distances in Cartesian space, it is straightforward to introduce extensions of the model using any typed geographic data. 
For example, operators may allow for low battery charge if the distance to an emergency landing site is low enough to arrive with remaining energy. 
Each point in the landscape incorporates two predictions based on the distribution of the battery charge per flight distance: one for the arrival at the target location and another for the subsequent arrival at an available emergency landing site.

For each location and type of map feature, both a distributional clause of the distance and the probability of the feature occupying the space are computed.
The object referencing the type of map feature in the HPLP will match the relational system underlying the employed GIS, in this case OSM.

%% file: figures/distributional_atoms.tex
\begin{figure*}
    \centering
    \begin{subfigure}{0.245\textwidth}
        \includegraphics[width=\textwidth]{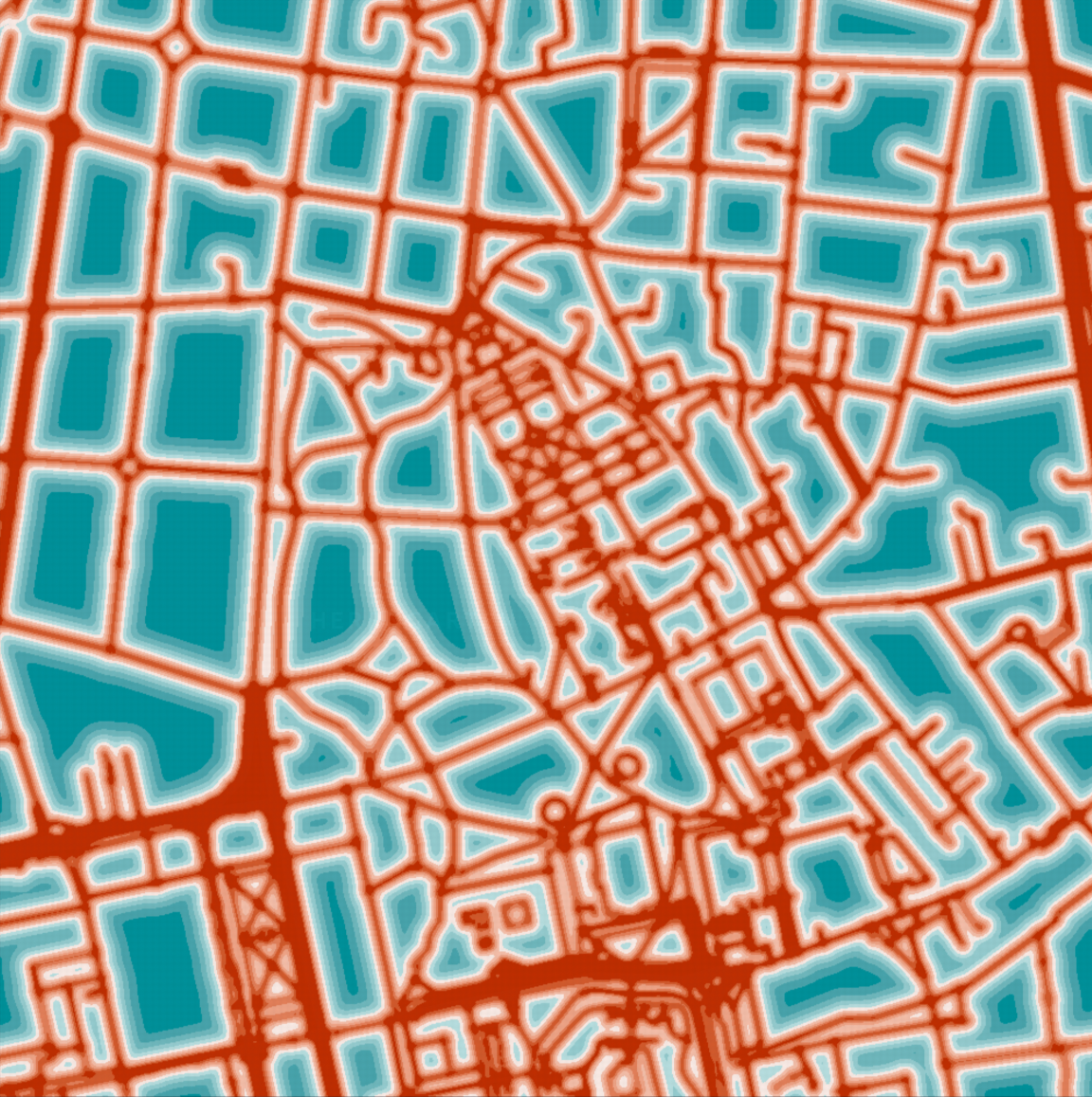}
        \caption{$\mu_D$}
    \end{subfigure}
    \begin{subfigure}{0.245\textwidth}
        \includegraphics[width=\textwidth]{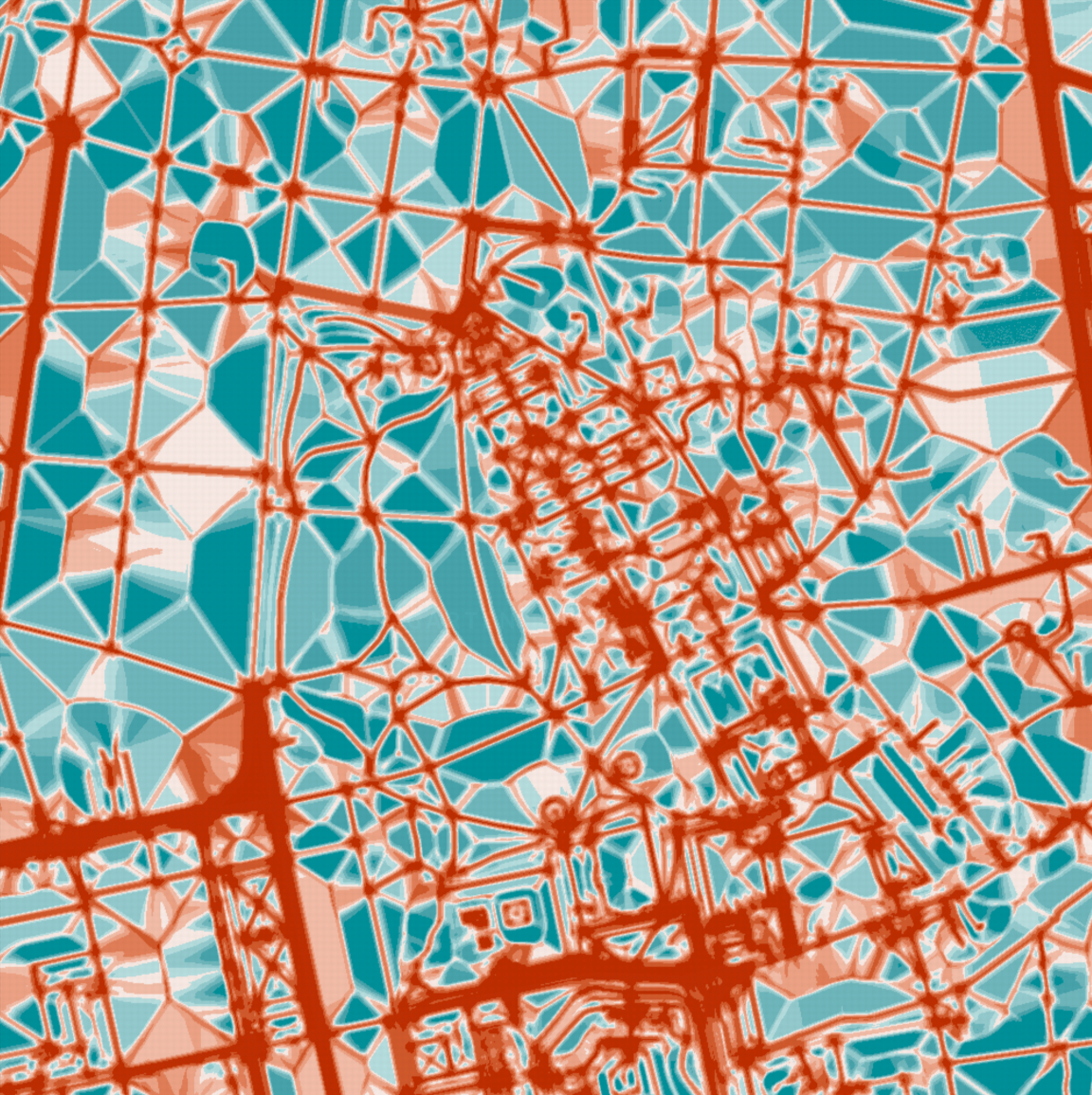}
        \caption{$\sigma^2_D$}
    \end{subfigure}
    \begin{subfigure}{0.245\textwidth}
        \includegraphics[width=\textwidth]{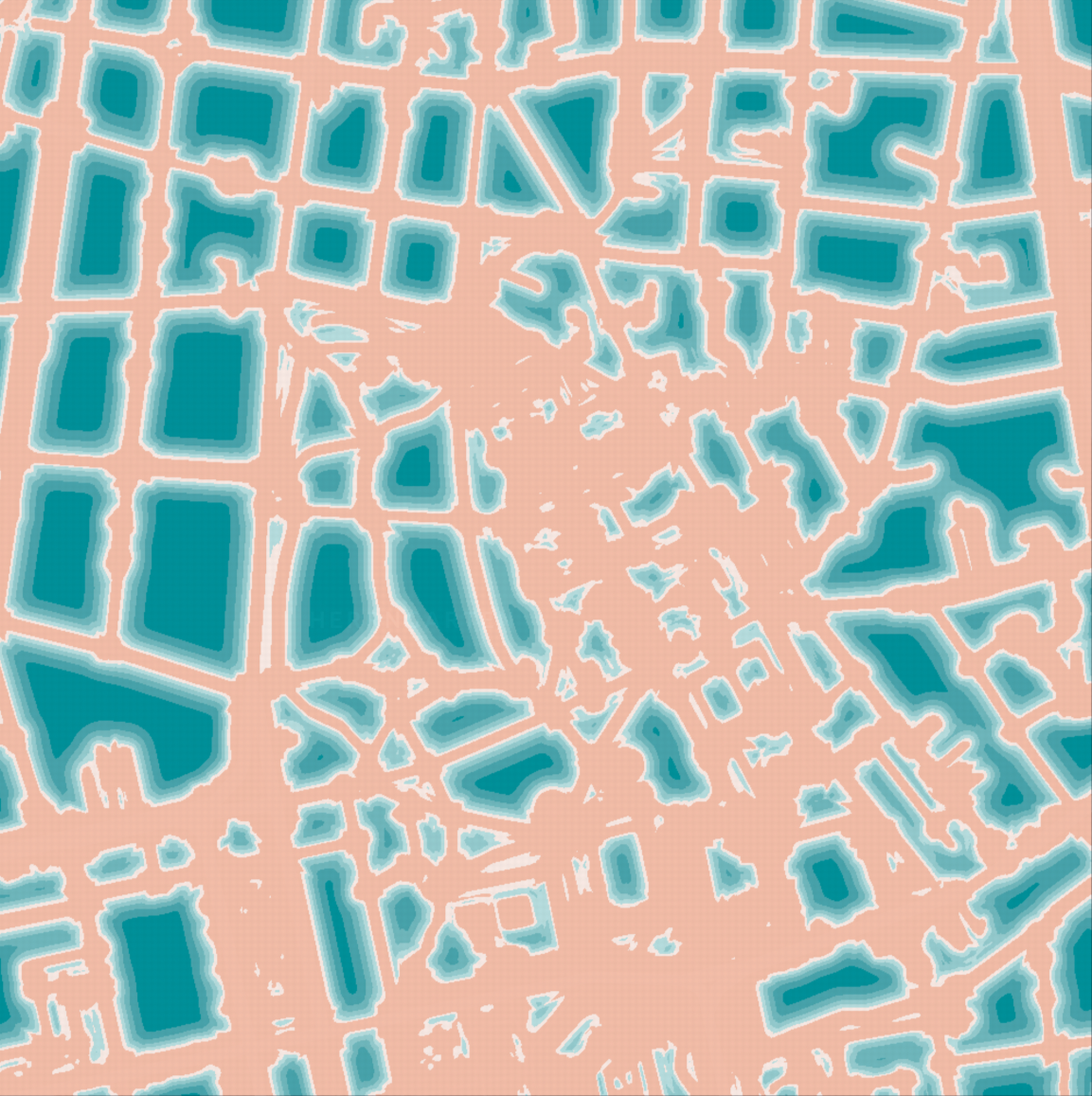}
        \caption{$P(D > 30)$}
    \end{subfigure}
    \begin{subfigure}{0.245\textwidth}
        \includegraphics[width=\textwidth]{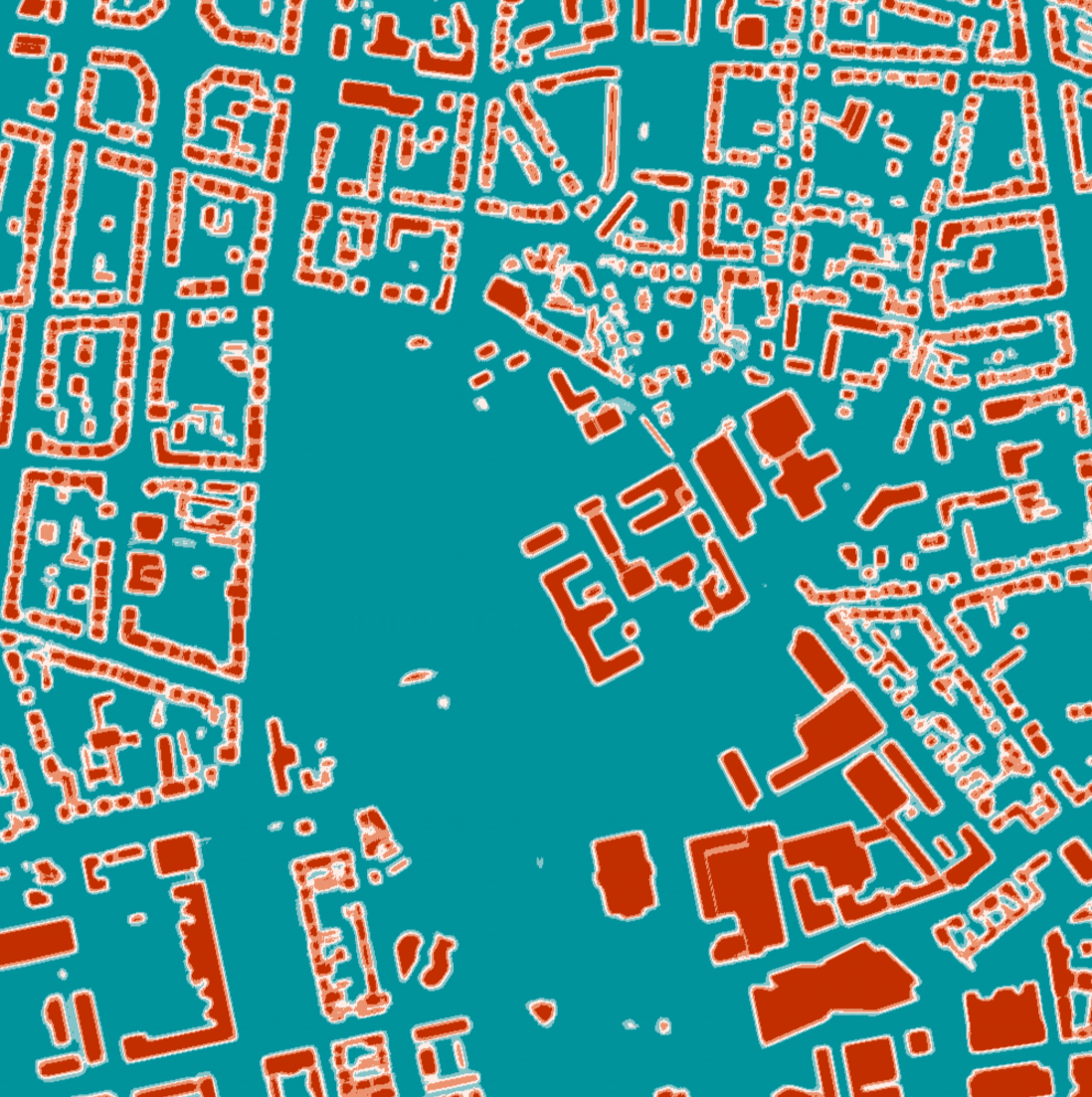}
        \caption{$P(O)$}
    \end{subfigure}
    \caption{
        \textbf{Probabilistic Clause Modules express uncertainties in geospatial data:}
        Here, a road network was queried from crowd-sourced data and random maps have been generated according to our uncertainty in the individual map features.
        For each location in the agent's navigation frame the parameters (a) and (b) of a normal distribution, modeling the distance to the closest road, are computed.
        From the cumulative distribution function of $\mathcal{N}(\mu, \sigma^2)$, or more generally via sampling, we can obtain the probability of a regulatory constraint being met, e.g., keeping a distance of over 30 meters (c).
        The color range transitions from red (low) to blue (high).
        Finally, (d) shows the probability of a location in the agent's navigation space being occupied by buildings.
        For visual clarity, the color range is inverse compared to the first three.
    }
    \label{fig:distributional_atoms}
\end{figure*}

%% file: listings/spatial_relations.tex
\begin{listing}[t]
    \centering
    \begin{minted}
    [
        frame=lines,
        autogobble,
        fontsize=\footnotesize
    ]{prolog}
        % Distributional clauses describing the distance of
        % each location and one of the types of map features
        distance(r0, c0, building) ~ normal(20, 0.5).
        distance(r1, c0, building) ~ normal(19, 0.4).
        ...
        
        % Probabilistic facts describing whether a location
        % is occupied by one of the types of map features 
        0.9::over(r0, c0, primary).
        0.8::over(r1, c0, primary).
        ...
    \end{minted}
    \caption{
        Probabilistic knowledge within the final HPLP $\mathcal{P}$ in the form of distributional, spatial atoms.
        Here, \textit{distance} and \textit{over} are generated relations and applied to each location in the agent's discretized navigation space.
        To keep the universe small, all features of a common type are aggregated into a single reference object, e.g., "building".
    }
    \label{listing:spatial_relations}
\end{listing}

%% file: listings/uam_model.tex
\begin{listing}
    \centering
    \begin{minted}
    [
        frame=lines,
        autogobble,
        fontsize=\footnotesize,
    ]{prolog}
    % Background knowledge and perception of drone
    registered.
    initial_charge ~ normal(90, 5).
    discharge ~ normal(-0.2, 0.1).
    weight ~ normal(2.0, 0.1).
    
    % Weather according to forecast
    1/10::fog; 9/10::clear.
    
    % Visual line of sight
    vlos(R, C) :- fog, distance(R, C, operator) < 250;
        clear, distance(R, C, operator) < 500.
    
    % Simplified OPEN flight category
    open(R, C) :- registered, vlos(R, C), weight < 25.
    
    % Charge to return to the operator
    can_return(R, C) :- B is initial_charge,
        O is discharge,
        D is distance(R, C, operator),
        0 < B + (2 * O * D).
    
    % Permit to fly in parks and close to major roads
    permit(R, C) :- over(R, C, park); 
        distance(R, C, primary) < 15;
        distance(R, C, secondary) < 10;
        distance(R, C, tertiary) < 5.
    
    % The Probabilistic Mission Landscape
    landscape(R, C) :- 
        permit(R, C), open(R, C), can_return(R, C).
    \end{minted}
    \caption{
        Simplified laws and constraints for the operation of an UAV.
        While the operator usually has to decide on-site whether and how they can maneuver their UAV, ProMis models and automates this decision.
        Probabilistic inference over the navigation space utilizing this model and extracted distributional knowledge of the environment yields a PML $\mathcal{L}$ for mission design.
    }
    \label{listing:uam_model}
\end{listing}

%% file: figures/scenarios.tex
\begin{figure*}
    \centering
    \begin{subfigure}{0.245\textwidth}
        \includegraphics[width=\textwidth]{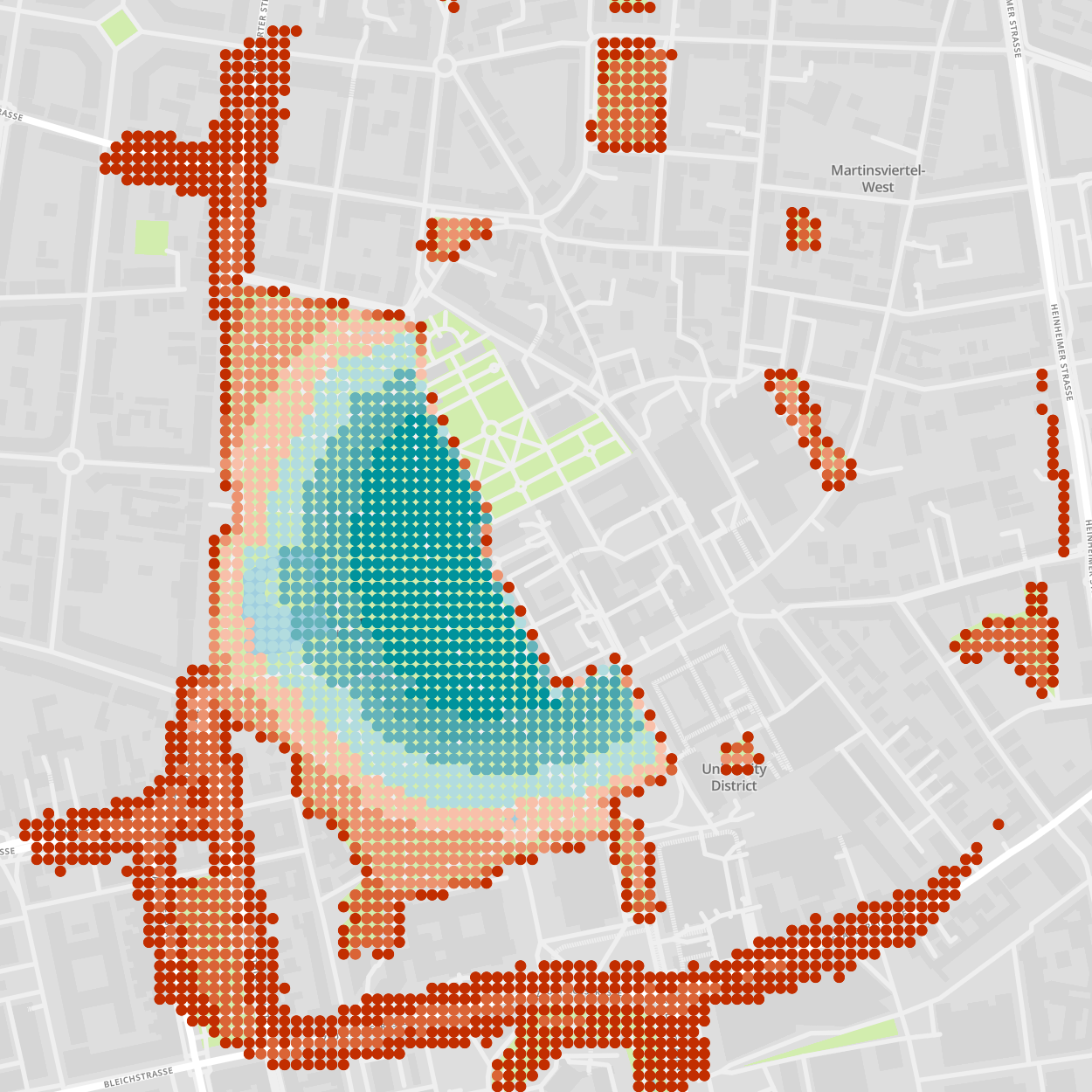}
        \caption{Park}
    \end{subfigure}
    \begin{subfigure}{0.245\textwidth}
        \includegraphics[width=\textwidth]{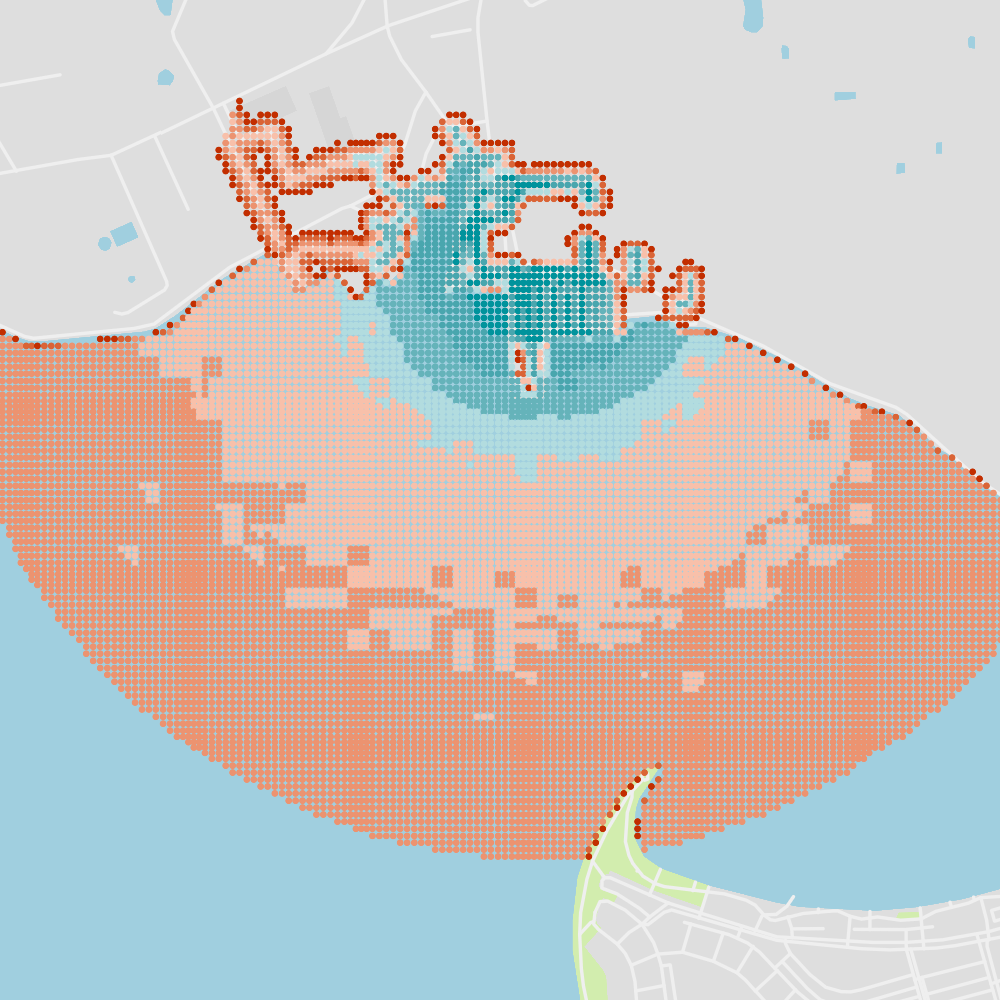}
        \caption{Bay Area}
    \end{subfigure}
    \begin{subfigure}{0.245\textwidth}
        \includegraphics[width=\textwidth]{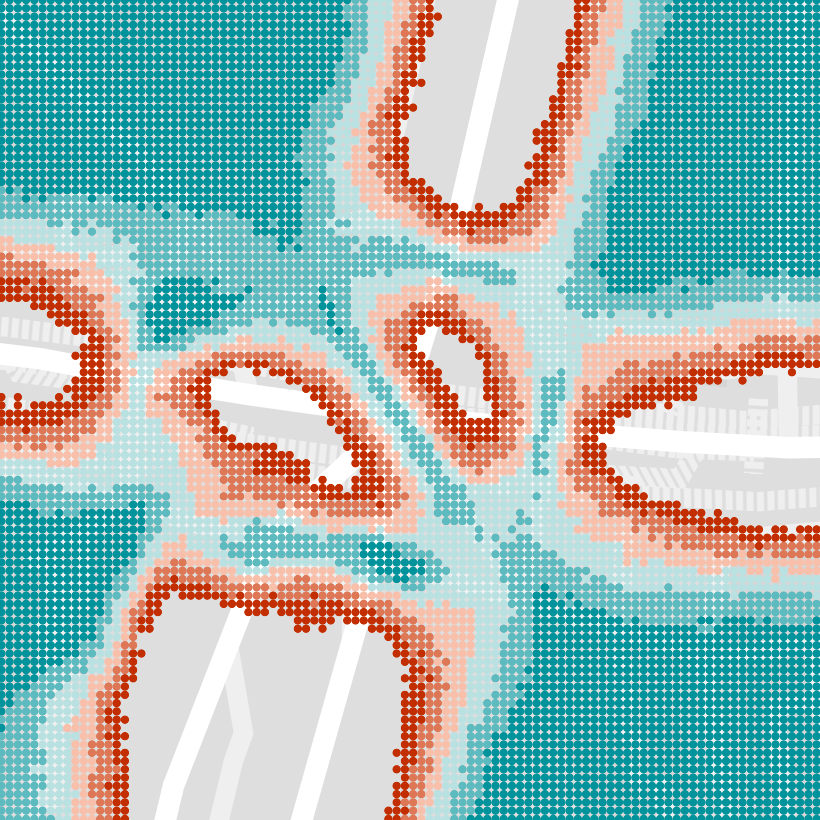}
        \caption{Crossing}
    \end{subfigure}
    \begin{subfigure}{0.245\textwidth}
        \includegraphics[width=\textwidth]{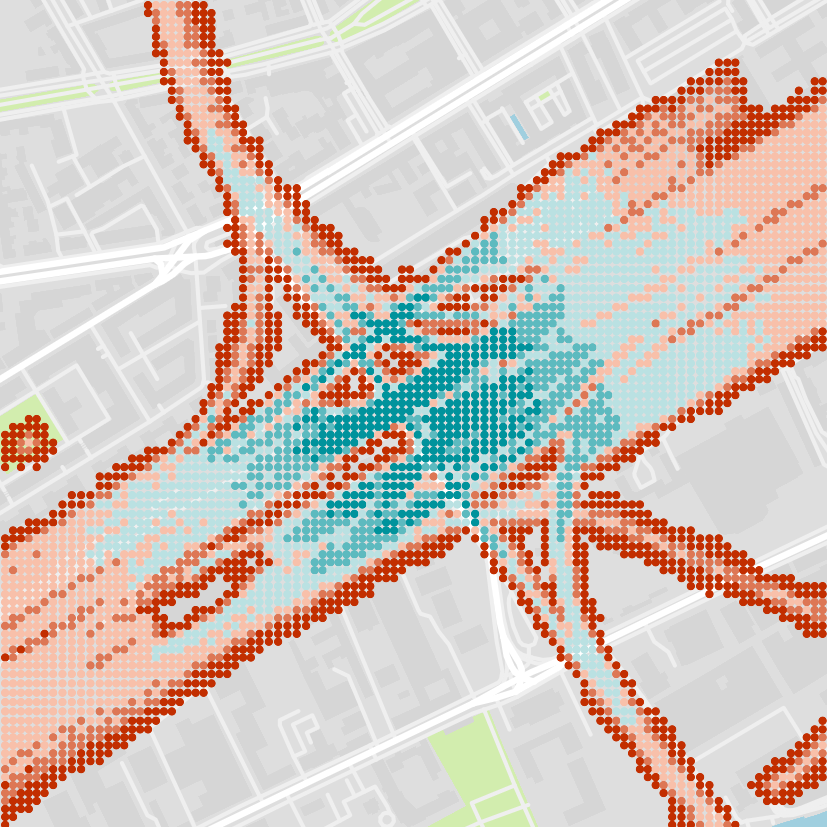}
        \caption{Rails}
    \end{subfigure}
    \caption{
        \textbf{ProMis generalizes to a variety of scenarios:}
        Here, we show the result of utilizing ProMis in (a) an urban city park with adjacent major road sections, (b) a bay area with increased visual-line-of-sight over open waters and accessible service roads and (c) a major junction with pedestrian crossings where a green signal allows crossing one from side to the other.
        Finally, (d) shows a scenario where the airspace right above passenger and cargo rails in the vicinity of the train station can be patrolled.
        High probabilities are represented by a dark cyan hue, turning red the lower the chance of fulfilling all constraints becomes.
        Values below 0.1 have been removed.
    }
    \label{fig:scenarios}
\end{figure*}

%% file: content/4_experiments.tex
\section{Experiments}
\label{sec:results}

\subsection{Probabilistic Missions in AAM Scenarios}

The aim of our research is to investigate the applicability of ProMis and the computational effort required for inference in HPLPs for mission design.
We will now present how ProMis supports UAV operations in a variety of scenarios.
Each scenario utilizes its own variation of mission design constraints, considering adjustments for the respective scenario and its relevant map features.
Regardless of the scenario, data was queried from OpenStreetMap using Overpass akin to Listing~\ref{listing:overpass_ql}.
All map features obtained this way have been assumed to be distributed according to a Gaussian translation $\vec{t} \sim \mathcal{N}(0, \text{diag}(10m, 10m))$ on their origin, representing the inherent uncertainty of crowd-sourced map data as discussed in Section~\ref{sec:methods}.

We will now discuss how ProMis can be applied to a diverse set of scenarios as can be seen in Figure~\ref{fig:scenarios}.
The first of the presented scenarios is driven by the model shown in Listing~\ref{listing:uam_model}.
It represents a simplified model for obtaining clearance to operate a UAV in the confinements of a public park and nearby major road segments while maintaining visual-line-of-sight and sufficient battery charge.
Hereby, a permission for free flight over the park area as well as in the vicinity of major roads is assumed.
Second, we apply ProMis to a Bay Area where we augment the definition of visual-line-of-sight (VLOS) to be extended over open waters if the weather is clear.
This shows how relating rules to the GIS provided typing system can conclude different behaviors based on the underlying map and uncertain weather conditions. 
Third, the scenario is moved to a crossing in a metropolitan environment where the UAV is discouraged from entering the airspace above the roads.
An exception is made for crossings while the traffic light shows green, showing how local perception or connections to other traffic systems may be leveraged.
Finally, a scenario in which the UAV can fly over a network of rails used for passenger and cargo transport is shown.
Here, the UAV is strictly confined to keep a narrow horizontal clearance to the rails while not straying away too far from its base-station.

ProMis reduces the need for operators to intuitively assess flight conditions, while remaining interpretable and adaptable by non-experts through changes to the constraints.
For distributional clauses, arithmetic clauses are resolved via sampling, which eliminates the need for definitive values for each aspect of the drone's operation. 
For example, a definitive weight rating is not necessary, allowing a delivery drone to change flight class swiftly as its weight changes, such as when delivering a package.
This enables mission design where a drone can carry a package to its destination under one set of restrictions and return to its station under another set.

\input{figures/time}

\subsection{Scaling Inference by Tiling the Navigation Space}

To assess the suitability of PML for mission design, we need to analyze the time taken to compute landscapes at different resolutions.
Without further strategy, the inference time will grow quadratically in the resolution.
This means that obtaining high resolution PMLs quickly becomes impractical even on a powerful computation device. 
Time measurements have been obtained on an Intel i7 9700K utilizing its 8 cores for multiprocessing.

By the way ProMis formulates background knowledge, the probability distributions of each location in the discretized navigation frame are independent of each other.
Therefore, we can make use of splitting up the inference task into multiple processes and thereby parallelize the task and cut down on computation time while retaining exact inference.
Here, we will split up the navigation area once horizontally and vertically with each tiling step, meaning for a tiling factor $s$, i.e., splitting the area $s$-times, we obtain $4^s$ smaller models on which inference can be run simultaneously.
In Figure~\ref{fig:time} one can see how this tiling enables us to scale ProMis to higher resolutions.

Tiling is an important consideration and the tiling factor a parameter with critical impact on computation time.
However, note that tiling might also be used in order to distribute work over time or devices.
More specifically, one might prioritize to compute tiles close to the agent first and queue up the rest for a later inference cycle.
Of course, since tiling makes use of parallel computing, moving the inference algorithms to a Single Instruction Multiple Data architecture is a promising direction for future work.

\subsection{Approximation of Probabilistic Mission Landscapes}

High-resolution landscapes are essential for accurately representing the environment, while low-resolution landscapes can misrepresent the model. 
With more fine-granular sampling, the PML becomes clearer and more nuanced, revealing additional information that can be used by the operator or planning method. 
However, it is challenging to determine the ideal resolution required to capture all critical information in complex models.
While Figure~\ref{fig:time} has shown how the time demand for landscape resolution can be tackled with a repeated tiling of the inference area, it is important to discuss the impact of choosing higher or lower resolutions in the first place.
Depending on the lateral and longitudinal extends of the inference area and the frequency of spatial complexity in the map, the resolution can be chosen lower without losing much accuracy.

We will now analyze this loss of accuracy when compared to a reference PML computed in a high resolution in the previously shown scenarios of Figure~\ref{fig:scenarios}.
We do so by running the knowledge extraction from the GIS and subsequent inference starting at a low resolution that is then iteratively increased.
Each PML we obtain this way is linearly interpolated and compared to the reference PML.
Figure~\ref{fig:error} shows the results for each scenario individually.
A trade-off between landscape quality and runtime needs to be considered here.
For instance, delaying the start of a mission to investigate the potential task-space may be appropriate for a user, but infeasible if quick decisions need to be made by the UAV itself. 
Still, further performance improvements might be crucial for iterative mission design processes where the landscape needs to be recomputed for each step in a mission.

\input{figures/error}

%% file: figures/time.tex
\begin{figure}
    \centering
    \includegraphics[width=\linewidth]{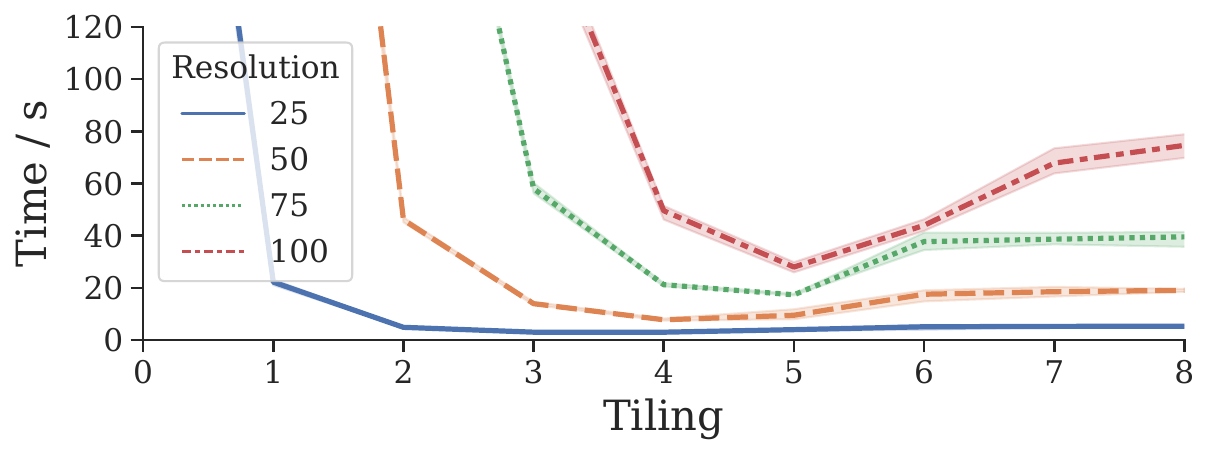}
    \caption{
        \textbf{Tiling reduces computation time:}
        Instead of creating a model of the entire navigation space and querying for all locations at once, we can divide the inference task repeatedly.
        Splitting the task decreases the computational cost, however, repeating this step too often, due to the cost of managing processes, actually increases the overall inference time.
        Here, the horizontal and vertical resolutions are equal.
    }
    \label{fig:time}
\end{figure}

%% file: figures/error.tex
\begin{figure}
    \centering
    \includegraphics[width=\linewidth]{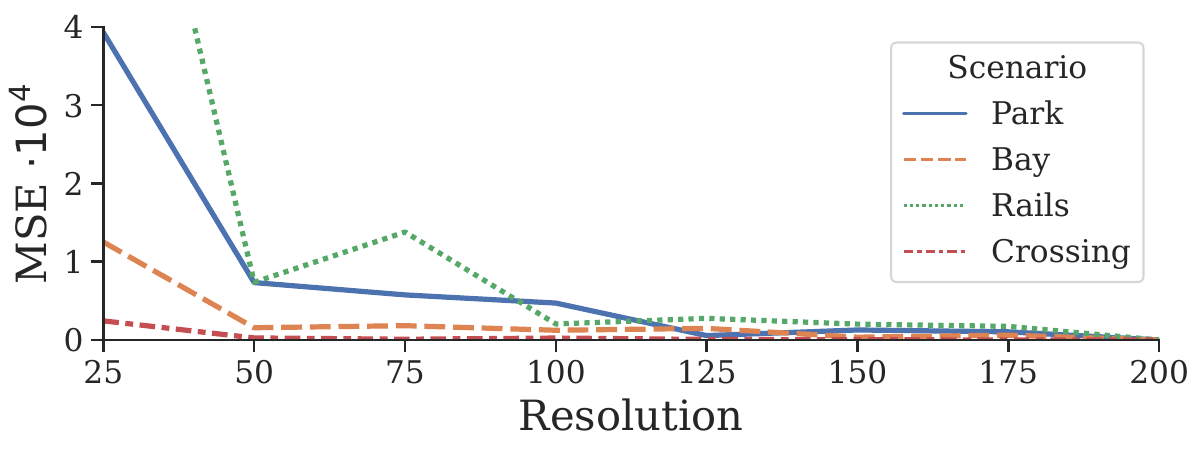}
    \caption{
        \textbf{Interpolation of low-resolution PMLs:}
        Here, we see the Mean Squared Error (MSE) when creating low resolution PMLs and, using bilinear interpolation, comparing to a $200 \times 200$ reference PML.
        As it depends on the spatial entropy, i.e., how much map features change, and the modelled laws and constraints, the disadvantage of using a low resolution inference area is hard to predict.
    }
    \label{fig:error}
\end{figure}

%% file: content/5_conclusion.tex
\section{Conclusion}
\label{sec:conclusion}

We have presented ProMis, an architecture for Probabilistic Mission Design based on Hybrid Probabilistic Logic Programs. 
ProMis enables adaptable and interpretable mission design by encoding static and dynamic, e.g., map and perception-based, knowledge into probabilistic and declarative logical expressions.
The hybrid probabilistic nature of the underlying inference mechanisms allows for the consideration of both discrete and continuous distributional knowledge, making it suitable for real-world scenarios.
Legal and operator constraints can be formulated in dependence of the derived probabilistic clauses for which we have presented two exemplary cases, \textit{distance} and \textit{over}.

Furthermore, we have shown the concept of Probabilistic Mission Landscapes, which combine models of the law, perception, and GIS data into a probabilistic view of the agent's navigation space.  
With this development, operators can better understand the legal situation regarding the deployment of their vehicle.
Using the landscape as the basis for motion planning can directly guide the search of both legal and safe trajectories.
Probabilistic Mission Landscapes not only impose the physical boundaries of static and dynamic entities for collision avoidance but also logically derived boundaries of the agent's motion due to legal and operator constraints.

However, we acknowledge that the complexity and spatial accuracy trade-off is a limitation for real-time decision-making and subject to future research.
For example, employing function approximation or interpolation methods to represent the Probabilistic Mission Landscape could further cut down on computation cost.
Scattering the locations to which ProMis is applied rather than inference on strict grids has the potential to reduce unnecessary work.
To this end, choosing inference locations based on the employed logic or the local entropy of the underlying map data is a promising direction.
To enable a richer environment to build logical constraints in, extensions to the Probabilistic Clause Modules are necessary.
In a similar vein, connecting ProMis with Open Universe Probabilistic Models is an important next step to design missions around, e.g., bystanders or other traffic participants.

Nevertheless, we have shown that ProMis intertwines important stakeholders and the autonomous agent, serving transparency and adaptation to human input while providing inference on mission-related, uncertain values.
Overall, we believe that ProMis provides an effective foundation for Advanced Aerial Mobility where low-level perception and high-level rules go hand in hand.

%% file: content/6_acknowledgment.tex
\section*{Acknowledgments}

Map data \copyright~OpenStreetMap contributors, licensed under the Open Database License (ODbL) and available from https://www.openstreetmap.org.
Map styles \copyright~Mapbox, licensed under the Creative Commons Attribution 3.0 License (CC BY 3.0) and available from https://github.com/mapbox/mapbox-gl-styles.